\documentclass[number]{elsarticle_arXvi}

\usepackage{amsmath,amsthm,amsfonts}
\usepackage{url}
\usepackage{fullpage}
\usepackage{amssymb}
\usepackage{multicol}
\usepackage{latexsym}
\usepackage{color}
\usepackage{natbib}
\usepackage{enumitem}
\usepackage{setspace}
\usepackage{lineno}
\usepackage{ifpdf}
\usepackage{xcolor}
\usepackage[section] {placeins}
\usepackage{pdflscape}
\usepackage{caption} 
\usepackage{graphicx}
\biboptions{sort&compress}

\begin{document}
\begin{frontmatter}
\title{\bf Ensemble random forest filter: An alternative to the ensemble Kalman filter for inverse modeling}

\author{Vanessa A. Godoy\corref{cor}} 
\author{Gian F. Napa-Garc\'ia} 
\author{J. Jaime G\'{o}mez-Hern\'{a}ndez} 
\address{Research Institute of Water and Environmental Engineering, Universitat Polit\`{e}cnica de Val\`{e}ncia, 46022, Valencia, Spain}

\cortext[cor]{Corresponding author: vaalde1@iiama.upv.es (Vanessa A. Godoy)}


\begin{abstract}
The ensemble random forest filter (ERFF) is presented as an alternative to the ensemble Kalman filter (EnKF) for the purpose of inverse modeling. The EnKF is a data assimilation approach that forecasts and updates parameter estimates sequentially in time as observations are being collected. The updating step is based on the experimental covariances computed from an ensemble of realizations and the updates are given as linear combinations of the differences between observations and forecasted system state values. The ERFF replaces the linear combination in the update step with a non-linear function represented by a random forest. In this way, the non-linear relationships between the parameters to be updated and the observations can be captured and a better update produced. The ERFF is demonstrated for the purpose of log-conductivity identification from piezometric head observations in a number of scenarios with varying degrees of heterogeneity (log-conductivity variances going from 1 up to 6.25 (ln m/d)$^2$), number of realizations in the ensemble (50 or 100), and number of piezometric head observations (18 or 36). In all scenarios, the ERFF works well, being able to reconstruct the log-conductivity spatial heterogeneity while matching the observed piezometric heads at selected control points. For benchmarking purposes the ERFF is compared to the restart EnKF to find that the ERFF is superior to the EnKF for the number of ensemble realizations used (small in typical EnKF applications). Only when the number of realizations grows to 500, the restart EnKF is able to match the performance of the ERFF, albeit at triple the computational cost.
\end{abstract}


\end{frontmatter}
\section{Introduction}
Characterization of the subsurface heterogeneity is of critical concern for modeling groundwater flow \cite[i.e.,][]{capilla-etal-99,li2011comparative,feyen2003bayesian,fernandez2007impact}  since it requires heterogeneous values of hydrogeologic parameters, which commonly are only sparsely available, if at all. To overcome the incomplete knowledge of the system and obtain better predictions with numerical models, state variables such as piezometric head---generally more extensively sampled---can be assimilated to improve the characterization of harder-to-measure parameters such as hydraulic conductivity \citep{carrera2005inverse,wen1999program}. Even with such an improvement, parameter heterogeneity is never completely known and its uncertainty also needs to be characterized. 

Stochastic data assimilation is an inverse modeling approach that can be used to characterize parameter heterogeneity and its uncertainty by assimilating state data sequentially in time \citep{zhou2014inverse}. 
The ensemble Kalman filter (EnKF) proposed by \citep {evensen1994sequential} is a very popular data assimilation (DA) method for stochastic inverse modeling that has been proven very efficient in numerous applications in fields as varied as atmospheric science, oceanography, geophysics, geotechnical and petroleum engineering, hydrology, or hydrogeology \citep{yin2015optimal, xu2016joint, shuai2016gathered, zhu2017data, chen2018joint, liu2018sequential,gelsinari2020feasibility,kim2020wave,he2021impact}. 

Data assimilation, for the purpose of inverse modeling, as implemented by the EnKF and its variants, is based on two main steps, a forecast of system evolution followed by an update (or correction) of the parameters describing the system based on the discrepancy between predictions and observations (at a few locations). The updates are computed using linear combinations with the weights calculated using covariance functions in a manner very similar to the geostatistical interpolation technique of cokriging. Such a linear scheme is a drawback of the Kalman-based DA methods since it is optimal when the system evolves in time following a linear state equation, but when the system evolves non-linearly, the model is suboptimal (although its performance may be quite good as demonstrated by their successful applications). A typical example of an EnKF implementation in which the relationship between the parameters and the state is non-linear is in inverse groundwater modeling \citep{evensen1994sequential,xu2013power}. 

One of the reasons for the success of the EnKF is that the experimental covariances are computed from ensembles of realizations that contain parameter values and their corresponding predictions.  The size of the ensemble is critical in the computational cost of the implementation; it should be as small as possible, to save CPU time, but it should be as large as possible to obtain good experimental covariance estimates and prevent filter inbreeding, the appearance of spurious correlations and avoid filter divergence. (These problems could be mitigated for small ensemble sizes with covariance localization techniques \citep{chen2006data, todaro2019ensemble,xu2013power}.) 
 
\cite{chen2006data} studied the sensitivity of the EnKF to, among other factors, the ensemble size, and the choice of the initial ensemble, and they showed that prior knowledge of the underlying field, such as the structure of the covariance function, plays an important role in data assimilation. Besides that, they found that a correct estimation of uncertainty may require large ensemble sizes. The need for large ensemble sizes and good prior knowledge of the spatial variability of the field, the linear nature of the updating step, and its big computational cost call for new strategies to improve available DA ensemble methods. 

In the last years, machine learning and big data are permeating all ambits of science and technology. The easiness with which large amounts of data are acquired in real-time and the new approaches to process them to build data-based predictive models have given rise to a new paradigm in the treatment of information that is starting to be used in environmental and water resources studies \citep{asher2015review,sit2020comprehensive, tahmasebi2021special, mariethoz2021machine}. In groundwater modeling, machine learning algorithms have been used mainly to replace process-driven models with data-driven ones to predict piezometric heads or solute concentrations from ancillary variables. The justification is that the data-driven models are cheaper to run and may capture relationships that could escape a process-driven analysis \citep{knoll2019large,al2020spatial, nguyen2020combining,sachdeva2021comparison,an2021improved}. Although these algorithms have proven their ability to deal with a wide range of problems in groundwater, they are seldom used for stochastic inverse modeling purposes, and to the best of the author's knowledge, it has not yet been used as a data assimilation algorithm capable to replace the restart EnKF (r-EnKF) \citep{chen2018joint, xu2018simultaneous, xu2016joint}. Without trying to be exhaustive, some example applications of machine learning in groundwater inverse modeling are the works by \cite{mo2019deep}, who combined an autoregressive neural network-based surrogate method for the forward model with an iterative local updating ensemble smoother (ILUES) \citep{zhang2018iterative} to solve high-dimensional contaminant transport inverse problems; \cite{bao2020coupling,bao2022variational}, who used Generative Adversarial Networks (GAN) \citep{goodfellow2014generative} to reparameterize hydraulic conductivity, using a low dimension latent variable, and then coupled it to an ensemble smoother with multiple data assimilation (ES-MDA) \citep{emerick2013ensemble}; or \cite{zhang2020using}, who used deep learning to ameliorate the ensemble smoother, although their starting ensemble was built with good prior knowledge of the underlying hydraulic conductivity spatial heterogeneity. 

Since the weakest point of the Kalman-based DA methods is the linear updating step, which, as mentioned before, is equivalent to cokrige the perturbations of hydraulic conductivity from the deviations between predicted and observed piezometric heads, it is proposed to replace the covariance-based updating step by a random forest-based updating. Random forest updating should be able to capture the multipoint non-linear relationships between conductivities and piezometric heads. This new method is termed ensemble random forest filter (ERFF). The idea of using random forests \citep{breiman2001random} was inspired by the work by \cite{hengl2018random} in which the authors propose, as an alternative to kriging, a new framework for spatial interpolation using random forest, demonstrating that this approach is capable of capturing relationships that go beyond the linear correlation intrinsic to the covariance. The framework proposed by \cite{hengl2018random} pursues the (non-linear) interpolation of an attribute from sparsely observed attribute values. In ERFF, however, the task is to interpolate piezometric head deviations (between observed and predicted values) to provide correction increments for hydraulic conductivity over the entire aquifer model. By taking advantage of the ensemble of realizations, and subtracting two by two all of them, a random forest can be trained to establish a relationship between differences in piezometric heads and differences in hydraulic conductivity. This procedure allows the generation of a large training data set with a relatively small number of realizations, solving also another problem commonly associated with the EnKF, which is the need for large ensemble sizes for the filter to be stable and good performing. Finally, the ERFF replaces the calculation and inversion of covariance matrices with random forest training. 

The ERFF is demonstrated in three synthetic aquifers of varying heterogeneity (variances ranging from 1.0 to 6.25 (ln (m/d))$^2$). A sensitivity analysis to the ensemble size and to the number of observations is carried out. Differently from previous researchers \citep{mo2019deep,goodfellow2014generative,zhang2020using} and in line with the work by \cite{xu2013power}, it is assumed that there is no prior information about the spatial heterogeneity of hydraulic conductivity, but only information about its mean value and its variance. \cite{xu2013power} have already shown the power of transient piezometric head in the characterization of hydraulic conductivity by the EnKF when no prior information is available. As will be shown, this power is intrinsic and can be taken advantage of by the ERFF, too. The concept of localization \citep{xu2013power,todaro2019ensemble} is also included in the implementation of the ERFF to reinforce the notion of spatial correlation by training the random forest giving more weight to the observations that are closer to the point being updated. The ERFF results are benchmarked against the r-EnKF.

The structure of this paper is as follows. First, the basics of ensemble Kalman filtering are introduced followed by the description of how the EnKF becomes the ERFF. Second, the three reference synthetic transient groundwater flow problems are described, together with the scenarios that will be analyzed. Third, the results for the different scenarios are shown, and a comparison of one of the scenarios with the r-EnKF is made. And fourth, the paper ends with a summary and an outlook on potential lines of continuing research.
 
\section{Stochastic data assimilation}
 
The EnKF algorithm \citep{evensen1994sequential} is the evolution of the Kalman filter \citep{kalman1960new} to handle nonlinear state transfer functions by using a Monte-Carlo approach. The EnKF (in the context of inverse modeling) is a sequential data assimilation method that updates the parameters of the model based on the discrepancies between model predictions and experimental observations. The relationship between parameters and observations must be known and a forward model relating parameters and state variables must be available. In the original implementation of the EnKF for inverse modeling, both model parameters and system states were updated, but it was found that the updated states might violate constitutive relationships (such as mass conservation) and the restart EnKF was introduced, whereby only model parameters are updated and the forecast for the next time step is always performed from time zero. The reader interested in the details of the EnKF is referred to the many papers published, particularly those by \citet{evensen1994sequential,evensen2003ensemble}. In the following, a brief description of the r-EnKF is presented that serves to introduce the ERRF.  

\subsection{r-EnKF: Ensemble data assimilation with covariance-based updating}

Consider a transient groundwater flow model in which piezometric heads are predicted based on the values of hydraulic conductivity on a discretized aquifer (plus corresponding boundary and initial conditions, and forcing terms). The forward model relating them is 
\begin{equation} \label{Eq:equation 1}
\mathbf{y}(t)=g(\mathbf{x},\mathbf{y}(t-\Delta t)),
\end{equation}
where $t$ is time, $\mathbf{x} \in \mathbb{R}^{n_p}$ is the hydraulic conductivity, $\mathbf{y} \in \mathbb{R}^{n_o}$ is the predicted system state at measurement locations, $g(\cdot)$ is a function that includes the numerical flow model plus an observation operator that extracts the predictions at observation locations,  ${n_p}$ is the number of cells in which the aquifer has been discretized and for which the hydraulic conductivity needs to be known in order to solve the numerical flow equation, and ${n_o}$ are the number of piezometric head observation locations. 
Piezometric heads are collected sequentially in time, and the purpose of the r-EnKF is, after each data collection, to update the hydraulic conductivities so that after a sufficient number of updates the hydraulic conductivity spatial distribution resembles the true but unknown one. The r-EnKF consists of an initialization step followed by repeated forecast and update steps as follows:
\begin{enumerate}
\item Initialization step. An initial ensemble of $n_e$ realizations of hydraulic conductivity $\mathbf{X}^{ini}$ is generated using statistical or geostatistical methods and incorporating as much prior knowledge as possible. In this paper, it is assumed that no prior information about the spatial variability of hydraulic conductivity is available, and the initial set of realizations is made up of homogeneous realizations, each one with a value drawn from a univariate distribution. 
\item Forecast step from time zero. In this step, the transient groundwater flow forward model is solved for each realization $i$ to obtain model predictions of the piezometric heads at time step $t$ using the latest update of the conductivities (for the first update, the initial ensemble of conductivities is used) starting from time zero. (Recall that to ensure that mass conservation is not violated by the piezometric heads at time $t$, the simulation is always restarted from time zero.)
\begin{equation} \label{Eq:equation 2}
\mathbf{y}_{i,t}=g(\mathbf{x}_{i,t-1},\mathbf{y}_{i,0}),\qquad i=1,\ldots,n_e,
\end{equation}
where $\mathbf{y}_{i,t}$ is the vector of forecasted piezometric heads at the $t^{\rm th}$ time step, and $\mathbf{x}_{i,t-1}$ is the last update of hydraulic conductivities at the previous time step $(t-1)$. For the first time step, $\mathbf{x}_{i,t-1}$ is $\mathbf{x}_i^{ini}$.  
\item Update step. The vector of hydraulic conductivities is updated based on the discrepancies between forecasted and observed piezometric heads. The updated parameter vector $\mathbf{x}^{u}$ is given, for the $i^{\rm th}$ realization at the $t^{\rm th}$ time step, by
\begin{equation} \label{Eq:equation 4}
\mathbf{x}^{u}_{i,t}=\mathbf{x}^{f}_{i,t}+\mathbf{K}_t\left[\mathbf{y}^{o}_{t}+\mathbf{\varepsilon}^{o}_{i,t}-\mathbf{y}^{f}_{i,t}\right],
\end{equation}
where the subscripts $i$ and $t$ refer to a specific realization and time step, respectively; $\mathbf{x}^{f}_{i,t=1} = \mathbf{x}^{ini}_{i}$ and $\mathbf{x}^{f}_{i,t} = \mathbf{x}^{u}_{i, t-1}$, $\mathbf{y}^{f}_{i,t}$ is the vector of model predictions at observation locations; $\mathbf{y}^{o}_{t}$ is the vector of state values at observation locations; ${\varepsilon}^{o}_{i,t}$ is the vector of observation errors (the observations errors have zero mean and a covariance matrix $\mathbf{R}_{t}$); and $\mathbf{K}_t$ is the Kalman gain matrix, given by 
\begin{equation} \label{Eq:equation 5}
\mathbf{K}_{t}=\mathbf{C}^{t}_{XY}\left(\mathbf{C}^{t}_{YY}+\mathbf{R}_{t}\right)^{-1},
\end{equation}
where $\mathbf{C}^{t}_{YY}$ is the auto-covariance of the state variables and $\mathbf{C}^{t}_{XY}$ is the cross-covariance between parameters and state variables for the $t^{\rm th}$ time step, which are computed from the ensemble of realizations as
\begin{equation} \label{Eq:equation 6}
\mathbf{C}^{t}_{YY}=\frac{1}{n_{\mathrm{e}}-1} \sum_{\mathrm{i}=1}^{\mathrm{n}_{\mathrm{e}}}\left(\mathbf{y}_{i,t}-\overline{\mathbf{y}}_{t}\right)\left(\mathbf{y}_{i,t}-\overline{\mathbf{y}}_{t}\right)^{\mathrm{T}}, 
\end{equation}
\begin{equation} \label{Eq:equation 7}
\mathbf{C}^{t}_{XY}=\frac{1}{n_{\mathrm{e}}-1} \sum_{\mathrm{i}=1}^{\mathrm{n}_{\mathrm{e}}}\left(\mathbf{x}_{i,t}-\overline{\mathbf{x}}_{t}\right)\left(\mathbf{y}_{i,t}-\overline{\mathbf{x}}_{t}\right)^{\mathrm{T}}, 
\end{equation}
with $\overline{\mathbf{x}}$ and $\overline{\mathbf{y}}$ being the ensemble means of parameters and predictions, respectively. 
\item Back to the forecast step.
\end{enumerate}

In a problem where there are ${n}_{p}$ parameters (in our case, ${n}_{p}$ will be the number of cells in the numerical model) and ${n}_{o}$ observations, vectors $\mathbf{x}^{u}_{i,t}$ and $\mathbf{x}^{f}_{i,t}$ have sizes ${n}_{p} \times 1$, vectors $\mathbf{y}^{o}_t$, ${\varepsilon}^{o}_{i,t}$, and $\mathbf{y}^{f}_{i,t}$ have sizes ${n}_{o}\times 1$, the Kalman gain $\mathbf{K}_t$ and the covariance $\mathbf{C}^{t}_{XY}$ are matrices of size ${n}_{p} \times {n}_{o}$, and the matrices $\mathbf{C}^{t}_{YY}$ and $\mathbf{R}$ are of size ${n}_{o} \times {n}_{o}$. When the observation errors are modeled as uncorrelated, $\mathbf{R}_{t}$ is a diagonal matrix. In the covariance matrix calculations, ${\overline{\mathbf{x}}_{t}}$ is a column vector of size ${n}_{p}\times 1$ with the average values of each parameter computed through the realizations, $\overline{\mathbf{x}}_{t}=\frac{1}{{n}_{e}} \sum_{\mathrm{i}=1}^{\mathrm{n}_{\mathrm{e}}} {\mathbf{x}_{i,t}} $, and, similarly ${\overline{\mathbf{y}}_{t}}$ is a column vector of size ${n}_{o}\times 1$ with the average values of each state variable computed through the ensemble of realizations, $\overline{\mathbf{y}}_{t}=\frac{1}{{n}_{e}} \sum_{\mathrm{i}=1}^{\mathrm{n}_{\mathrm{e}}} {\mathbf{y}_{i,t}}$.

\subsection{ERFF: Ensemble data assimilation with random forest-based updating}

The ERFF proposal is to replace the linear updating in Eq.~\eqref{Eq:equation 4} with a non-linear update based on a random forest prediction. Eq.~\eqref{Eq:equation 4} can be rearranged as follows
\begin{equation} \label{Eq:equation 8}
\mathbf{x}^{u}_{i,t} - \mathbf{x}^{f}_{i,t}=\mathbf{K}_t\left[\mathbf{y}^{o}_{t}-\mathbf{y}^{f}_{i,t}+\mathbf{\varepsilon}^{o}_{i,t}\right],
\end{equation}
and rewritten as
\begin{equation} \label{Eq:equation 9}
{\Delta}\mathbf{x}_{i,t}=\varphi ({\Delta}\mathbf{y}_{i,t}),
\end{equation}
where ${\Delta}\mathbf{x}_{i,t}$ and ${\Delta}\mathbf{y}_{i,t}$ are the correction (to be applied to the current estimate of the parameters) and the discrepancy (between state predictions and observations)

vectors, respectively. In the r-EnKF, function  $\varphi$ is a linear combination of discrepancies, where $\mathbf{K}_{t}$  (Eq. (\ref{Eq:equation 5})) is the matrix of coefficients. In the ERFF, $\varphi$ will be replaced by a random forest regressor, which should be able to capture any linear or non-linear relationship existing between ${\Delta}\mathbf{x}_{i,t}$ and ${\Delta}\mathbf{y}_{i,t}$.

Random forest regression is a supervised machine learning algorithm for building a predictor ensemble with a set of decision trees (that is, a forest) that grow in bootstrapped sub-samples of the dataset (that is, randomly selected samples with replacement). Predictions are obtained by aggregating the various predictors from each decision tree into a single average value \citep{breiman2001random, cutler2012random, biau2012analysis}. The bootstrap aggregation procedure used in random forest produces robust and highly accurate predictions without overfitting \citep{biau2012analysis, hengl2018random}. As the mathematical framework of the random forest itself is not the focal point of this work, interested readers are encouraged to refer to \cite{breiman2001random}, \cite{cutler2012random}, and \cite{biau2012analysis} for a  more in-depth analysis of the technique. 

A random forest has to be built for each cell in the model where log-conductivity is to be estimated. Once built, the discrepancies between forecasted piezometric heads (different for each realization) and observed values are fed to the random forest to provide an estimate of the log-conductivity perturbation to apply, at that specific cell, to each realization. The ERFF consists of the same steps as the r-EnKF: an initialization step followed by repeated forecast and update steps. The difference lies in the update step, which is done using random forests as explained next.
Consider a set of $n_e$ realizations of the hydraulic conductivity and the associated $n_e$ realizations of the piezometric heads at a given time step. With such a set, subtracting two by two each conductivity realization and its associated piezometric heads, an ensemble of $n_e'=n_e(n_e-1)/2$ realizations of differences can be built

\begin{equation}
\left.
	\begin{matrix} \label{Eq:equation 10}
{\Delta \ln \mathbf K}_{i_3,t}&=&{\ln \mathbf K}_{i_2,t} - {\ln \mathbf K}_{i_1,t} \\
{\Delta \mathbf  h}_{i_3,t}&=&{\mathbf h}_{i_2,t} - {\mathbf h}_{i_1,t}
\end{matrix} \right\} i_1=1,\ldots,n_e-1, i_2< i_1 \leq n_e , i_3=1,\ldots,n_e'
\end{equation}
where $\Delta \ln \mathbf K_{i_3,t}$ is a realization of log-conductivity differences at time step $t$, and $\Delta \mathbf h_{i_3,t}$ is a realization of piezometric head differences at the same time step and for the same conductivity realizations used to obtain the log-conductivity difference. Next, consider that observations have been taken at a subset of $n_o$ locations, these observations are going to depart from the forecasted values, and the differences between observations and forecasts will change for each realization of log-conductivity. Consider now a specific cell in the numerical model, $j$; from the ensemble of differences it is possible to build a training data set composed of 
\begin{equation}
	\left. 
	\begin{matrix}
		\Delta \ln K_{i,j,t}\\
		\Delta h_{i,k,t}, k=1,\ldots,n_o
	\end{matrix}
	\right\}  i=1\ldots,n_e' 
\end{equation}
from which to train a random forest to predict the perturbation of log-conductivity at location $j$ associated with perturbations of the piezometric heads at the set of $n_o$ locations. Once this random forest is trained, the differences between the observed heads and the predicted ones in each realization are calculated and the random forest is used to predict a log-conductivity difference to apply to the current value of log-conductivity at that specific location. This procedure is repeated for each cell in the aquifer until all conductivity values are updated. 

In order to reinforce the need to account for spatial correlation, the head differences are weighted before their use according to
\begin{equation}  \label{Eq:equation 12}
	\Delta' h_{i,k,t} = \Delta h_{i,k,t} \lambda^{-1}({r})
\end{equation}
where $i$ is the realization index, $k$ is the observation index, $t$ is the time index, $r$ is the Euclidean distance between the observation and the point where log-conductivity has to be updated,  and $\lambda$ is a function that decays with $r$
\begin{equation}\label{Eq:equation 13}
	\lambda (r) = 1-\exp (-\frac{r^2}{3a})
\end{equation}
where $a$ is the distance beyond which no spatial correlation between the head differences and the log-conductivity differences is expected. The rationale of using Eq.~\eqref{Eq:equation 12} is the following: when the observation location is close to the log-conductivity location being updated, $\lambda$ is close to one and no correction is introduced, but when the head difference is far from the log-conductivity, the value of $\lambda$ is close to zero and the head difference is amplified in a way that the random forest will interpret that there is no relationship between head differences and log-conductivity differences. In this way, head differences close to the point being updated will receive larger weight in the log-conductivity update than head differences that are further apart.

The random forest was implemented using the scikit-learn library in Python \citep {pedregosa2011scikit}. Before running the different scenarios that will be described in the next section, it was necessary to tune the hyperparameters of the algorithm. This is probably the most tedious part of the ERFF, which is always subject to some subjective decisions. A number of preliminary runs were performed splitting the ensemble of differences into two subsets, 90\% for training and 10\% for validation, and a sensitivity analysis was performed to derive the best hyperparameter values. The values finally chosen were: number of trees in the forest 120,  minimum number of samples required to split an internal node  2,  minimum number of samples required to be at a leaf node  3,  number of features to consider  0.65, and random state  10. All other hyperparameters were set at their default values as defined in scikit-learn. 

\section{Synthetic examples}
\begin{figure}[htbp]
\begin{center}
\includegraphics*[scale=0.30, angle=0]{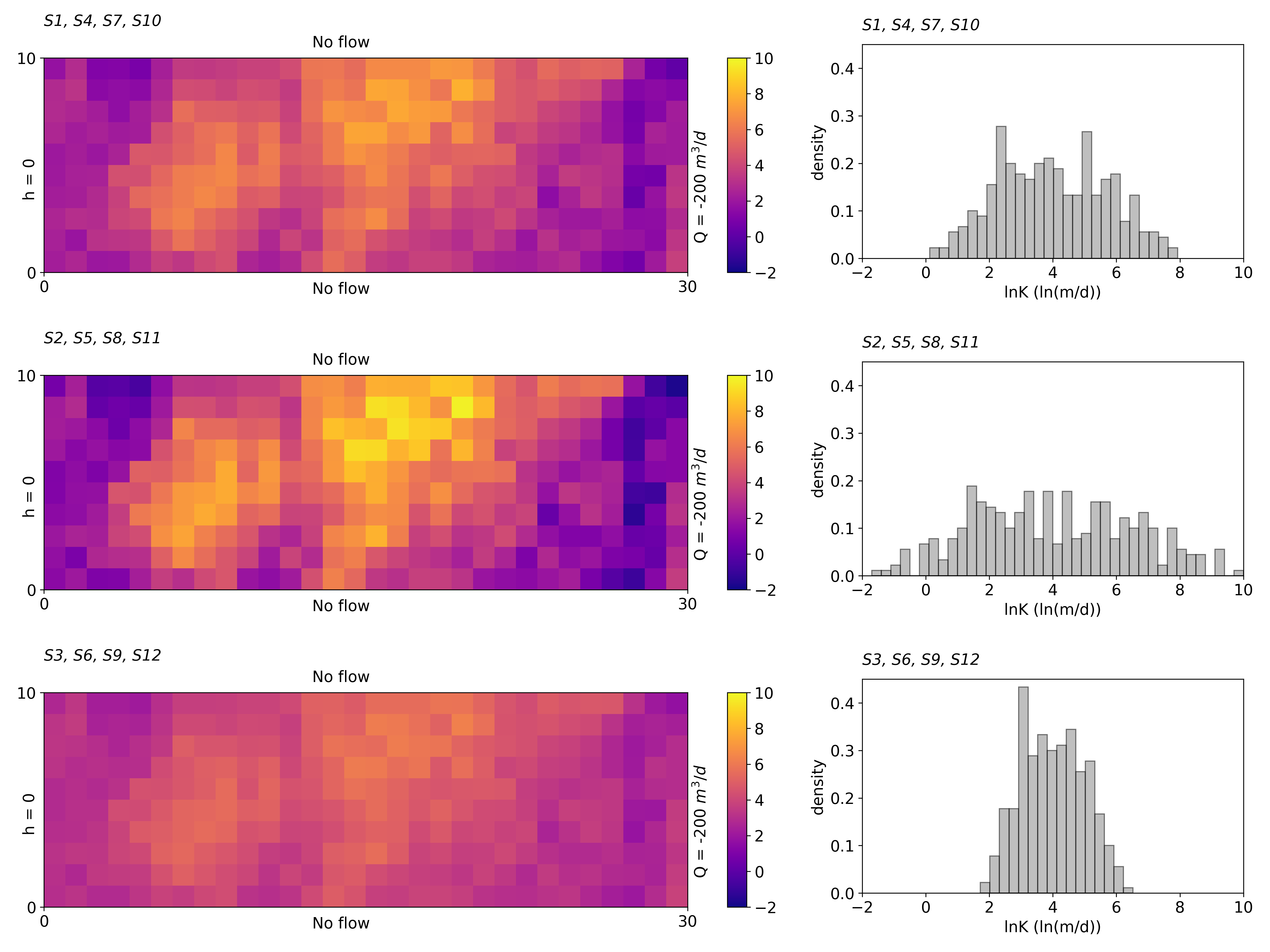}
\end{center}
\caption{Reference fields and corresponding histograms.}
\label{fig:Figure 1}
\end{figure}

Three synthetic two-dimensional, heterogeneous, and confined aquifers are built on a domain composed of 30 by 10 cells, each one being 1 m by 1 m.  The GCOSIM3D code \citep{gomez1993joint} was used to generate the three reference log-conductivity fields with standard deviations of 1.0, 1.7, and 2.5 ln (m/d), and all of them with a mean of 4.0 ln (m/d) and a spherical variogram with maximum and minimum ranges of 20 and 10 meters, respectively, with the direction of maximum continuity oriented at 30$^\circ$ counterclockwise with respect to the east axis. Transient groundwater flow is simulated in all three synthetic aquifers under the following conditions: north and south boundaries are impervious; along the east boundary, a flow of -200 m\textsuperscript{3}/d is prescribed; heads of 0 m are prescribed along the west boundary, and initial hydraulic heads are set to 0 m everywhere. Figure~\ref{fig:Figure 1} shows the three log-conductivity reference fields with indication of the groundwater flow boundary conditions, along with their histograms.  The total simulation time is 5 days discretized into 100 time steps.  Transient groundwater flow is numerically solved by MODFLOW 2005 \citep{harbaugh2005modflow} in FloPy \citep{bakker2016scripting}.

\begin{figure}[htbp]
\begin{center}
\includegraphics*[scale=0.3, angle=0]{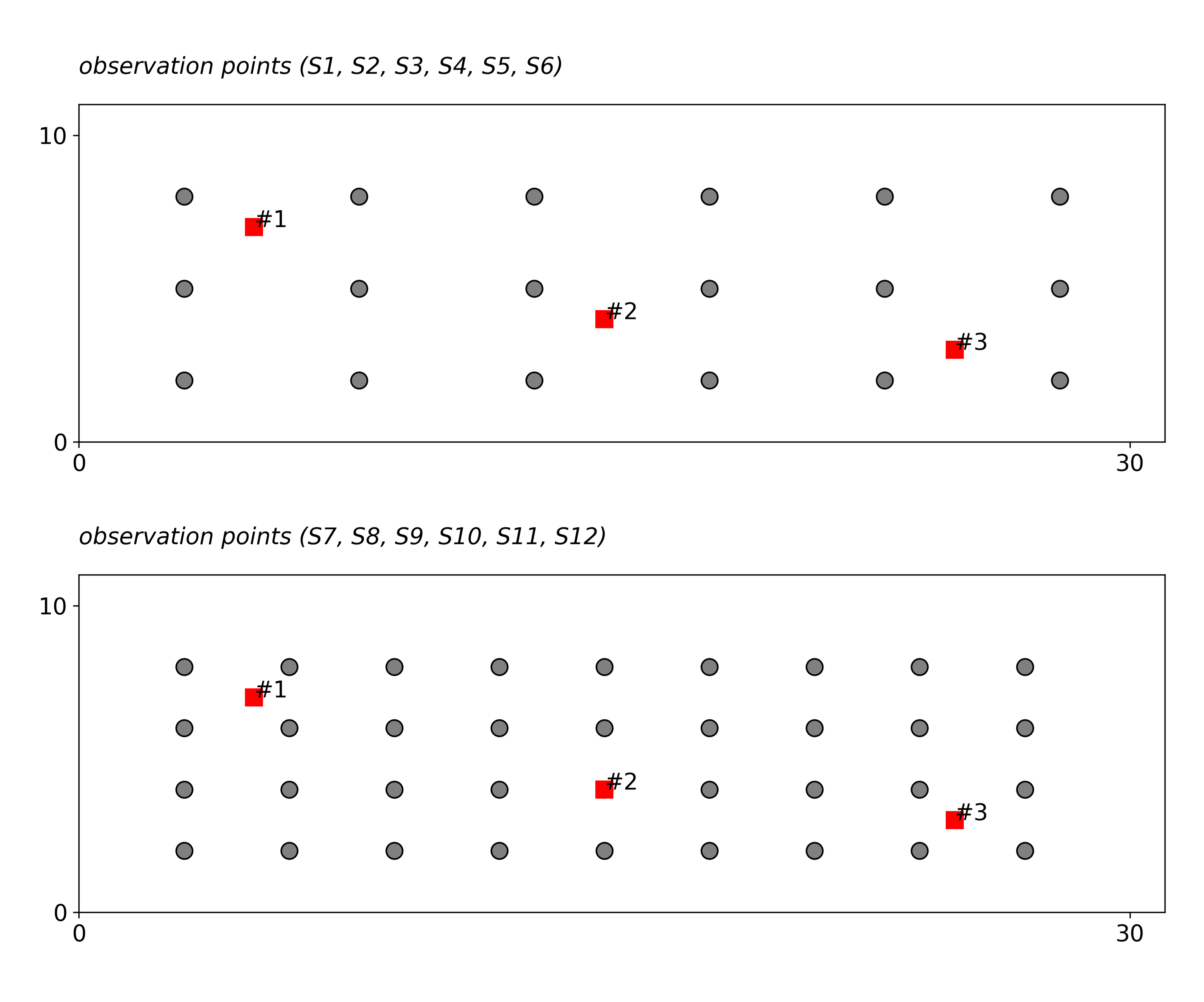}
\end{center}
\caption{Observat ion (circle) and control (square) points}
\label{fig:Figure 2}
\end{figure}

Each transient simulation for each reference field was sampled at the locations shown in Figure~\ref{fig:Figure 2}. The sampled values will be assimilated by the ERFF with the objective of retrieving the spatial heterogeneity of the reference fields. Only the observations for the 26 first time steps are used during the assimilation. The remaining 74 time steps are used for validation. Figure~\ref{fig:Figure 2} also shows three control points that will not be used during the assimilation but that will serve to validate the final results. 
Twelve scenarios were defined to evaluate the performance of the ERFF. The scenarios were built to analyze the influence of the number of realizations in the ensemble, the number of observation points, and the standard deviation of the reference field. Table~\ref{tab:Table1} summarizes the scenarios considered.

\begin{figure}[htbp]
\begin{center}
\includegraphics*[scale=0.3, angle=0]{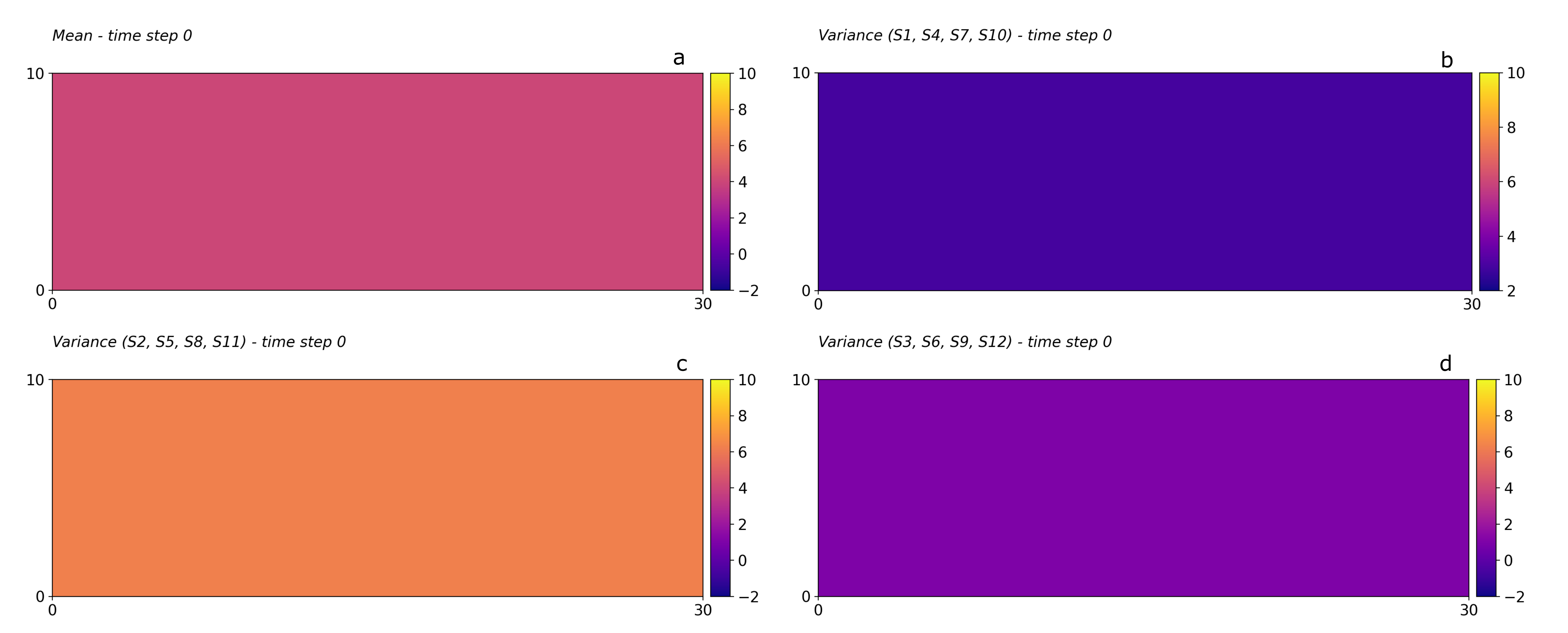}
\end{center}
\caption{Mean and variances of the initial ensembles.}
\label{fig:Figure 3}
\end{figure}

For the generation of the initial ensemble of realizations, it is assumed that no prior information about the spatial variability of conductivity is available. For this reason, the assimilation procedure for all scenarios starts with an ensemble of homogeneous log-conductivity realizations drawn from Gaussian probability distributions of mean 4.0 ln (m/d) and standard deviations of 1.0, 1.7, and 2.5 ln (m/d) according to the last column in Table~\ref{tab:Table1}. Figure \ref{fig:Figure 3} displays the ensemble means (Fig. \ref{fig:Figure 3}a) and the ensemble variances for all scenarios (Fig. \ref{fig:Figure 3}b-d) for the initial log-conductivity fields. As expected, these values are homogeneous and equal to the prior mean (the same for all scenarios) and variance (different for the scenarios according to Table~\ref{tab:Table1}). 

\begin{table}
	\centering
    \caption{Scenarios considered}
    \captionsetup[table]{skip=10 pt}
    \label{tab:Table1}   
    \begin{tabular}{c c c c}
     \\ 
     \hline
      \textbf{ Scenario} & \textbf{\# observations} & \textbf{ \# realizations} & \textbf{Standard deviation}\\   
      \hline
		S1 & 18 & 50 & 1.7\\
		S2 & 18 & 50 & 2.5\\
		S3 & 18 & 50 & 1.0\\
		S4 & 18 & 100 & 1.7\\
		S5 & 18 & 100 & 2.5\\
		S6 & 18 & 100 & 1.0\\
		S7 & 36 & 50 & 1.7\\
		S8 & 36 & 50 & 2.5\\
		S9 & 36 & 50 & 1.0\\
		S10 & 36 & 100 & 1.7\\
		S11 & 36 & 100 & 2.5\\
		S12 & 36 & 100 & 1.0\\
	\hline 
    \end{tabular}
\end{table}

Each scenario is used to study the ERFF for the identification of the reference field with the same standard deviation in the last column of Table~\ref{tab:Table1}. Apart from the standard deviation, the scenarios differ in the number of members of the initial ensemble, which can be 50 or 100, and the number of head observation points, which can be 18 or 36 as shown in Figure \ref{fig:Figure 2}. Hydraulic heads are collected and assimilated every time step for the first 26 time steps, then the model continues running until time step 100. 

In all scenarios, localization is used, with a parameter $a$ in \eqref{Eq:equation 13} equal to 12 m, implying that virtually no spatial correlation between head differences and log-conductivity differences exists beyond this distance.

Finally, for the sake of completeness, the r-EnKF was also applied to scenario S1 and used as a benchmark for ERRF.

\section{Results and discussion}

\begin{landscape}
\begin{figure}[htbp]
\begin{center}
\includegraphics*[scale=0.3, angle=0]{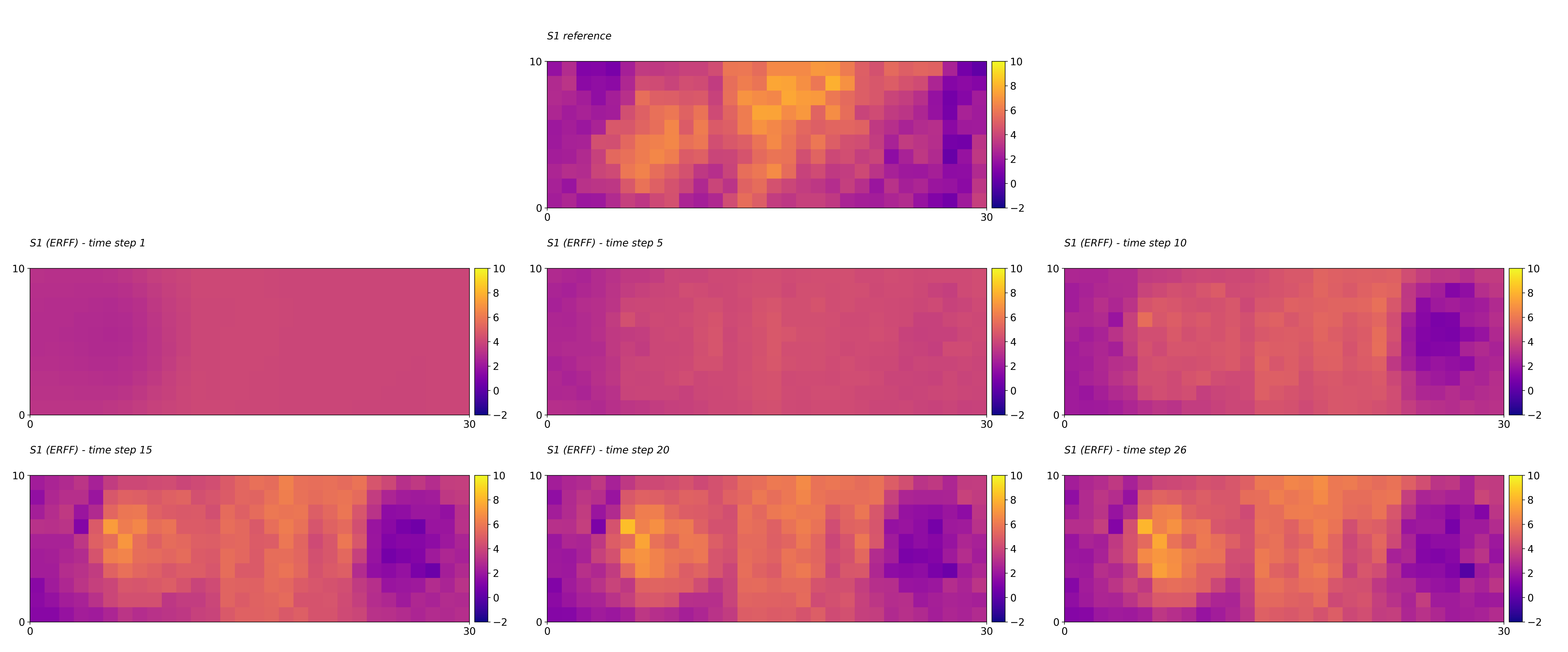}
\end{center}
\caption{Evolution in time of the ensemble mean for scenario S1.}
\label{fig:Figure 4}
\end{figure}
\end{landscape}

\begin{figure}[htbp]
\begin{center}
\includegraphics*[scale=0.3, angle=0]{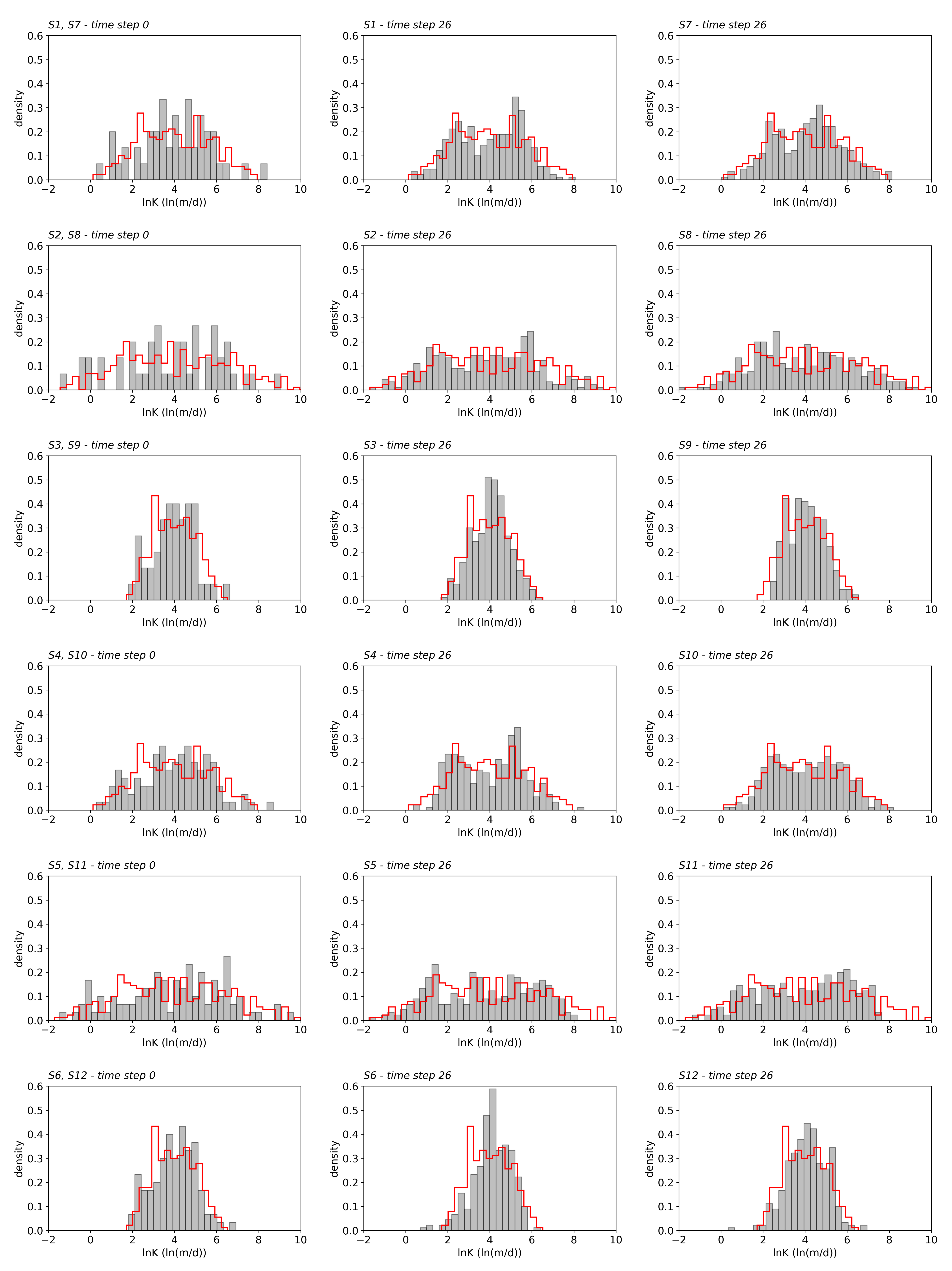}
\end{center}
\caption{Initial (left column) and final histograms of the final estimated log-conductivities by the mean of the ensemble at time step 26 (central column, scenarios with 18 observations; right column, scenarios with 36 observations). The hollow red histograms correspond to the reference fields.}
\label{fig:Figure 5}
\end{figure}

Figure \ref{fig:Figure 4} shows, for scenario S1, how the mean of the ensemble of realizations evolves as observations are assimilated. It can be observed how, starting from a homogeneous mean, heterogeneity is gradually introduced in the ensemble of realizations after each assimilation step, and, by step 26, the mean of the ensemble is a good estimate of the reference. The large-scale features of the reference are already visible in step 10, and, by step 20, the short-scale features are displayed, too; not many changes are noticeable after step 20. Similar time evolutions are observable in the rest of the scenarios, although not shown here. These results are very promising, particularly under the consideration that no prior information about spatial heterogeneity is used. Figure~\ref{fig:Figure 5} shows the evolution of the histograms of all realizations for each scenario. In the first column,  the histograms for all values in the initial ensembles of realizations for each scenario are shown as solid gray bars. In the second and third columns, the histograms of the updated fields are shown. In all three columns, the hollow red histogram is the histogram in the reference field. There is not much difference between the initial and the updated histograms, although it is clear that there is a shift towards a better fit to the reference histogram in the updated realizations; however, it is important to notice that the spatial heterogeneity of the realizations has gone from homogeneous values in each realization in the first column to heterogeneous ones trying to replicate the reference so as to match the observed piezometric heads in the other two columns. As already said, the only statistical information used for the generation of the initial ensemble is the probability distribution from which to draw the homogeneous values for each realization; these distributions were chosen to match the ones used to generate the references, but it can be said that starting from a uniform distribution with reasonable ranges will yield the same results, meaning that the method is capable of retrieving the spatial patterns of the heterogeneous log-conductivity field with virtually no prior information on this parameter. The difference between the scenarios in the second and third columns of Figure~\ref{fig:Figure 5} is the number of observation points, 18 and 36, respectively. With 36 observations, the final updated histograms are slightly closer to the reference ones. 

The performance of the method was further analyzed through sensitivity analysis to three variables: number of observation points, ensemble size, and hydraulic conductivity variance. 

Figure \ref{fig:Figure 6} shows the ensemble mean of the updated log-conductivity fields after the 26$^{\rm th}$ assimilation time step for all scenarios. The left column presents the final mean log-conductivity field corresponding to a standard deviation of 1.7 ln (m/d) while the center and right columns present the final fields corresponding to standard deviations of 2.5 and 1.0 ln (m/d), respectively. The first row presents the reference fields for comparison purposes, the second and third rows refer to scenarios with 18 observation points, and the fourth and fifth rows show the scenarios with 36 observation points. Figure \ref{fig:Figure 7} presents the ensemble variance of the updated log-conductivity fields after the 26$^{\rm th}$ assimilation time step for all scenarios. And Figure \ref{fig:Figure 8} shows the standardized discrepancy between the reference and the ensemble mean of the updated fields for each scenario computed as the difference between reference value and ensemble mean over the scenario standard deviation. (In the latter two figures, no reference row is displayed, the first and second rows correspond to the scenarios with 18 observation points, and the third and fourth rows scenarios with 36 observation points. Left, center, and right columns correspond to scenarios with log-conductivity standard deviations of 1.7, 2.5, and 1.0 ln (m/d), respectively.) From these three figures, one can qualitatively observe that the method successfully reproduces the heterogeneity of the reference fields regardless of the scenario. It is worth noting that the results are very similar independently of the number of simulations, the number of observations, or the standard deviation of the reference field. Only a slight improvement is found when the number of observations is doubled. Further analyses carried out and not presented here, showed that the number of observations could be reduced down to ten, and still, the ERFF was able to recover the heterogeneity of the underlying conductivity fields. The success of the approach must be related to the ability of random forests to extract non-linear relationships between explanatory variables (piezometric head differences) and the parameters (hydraulic conductivity differences). 

Figure \ref{fig:Figure 9} presents a comparison of the ERFF and the r-EnKF using the same number of observation points, ensemble size, and standard deviation of the reference field. The ERFF arrives to a more accurate mean field (see Fig. \ref{fig:Figure 9}(b)) than the r-EnKF (see Fig. \ref{fig:Figure 9}(c)). The better performance can be verified by comparing the variance, errors, and histograms of the final fields for both methods. This does not mean that the r-EnKF is disqualified for the purposes of inverse modeling, but under these settings, the ERFF is better. 

\begin{landscape}
\begin{figure}[htbp]
\begin{center}
\includegraphics*[scale=0.3, angle=0]{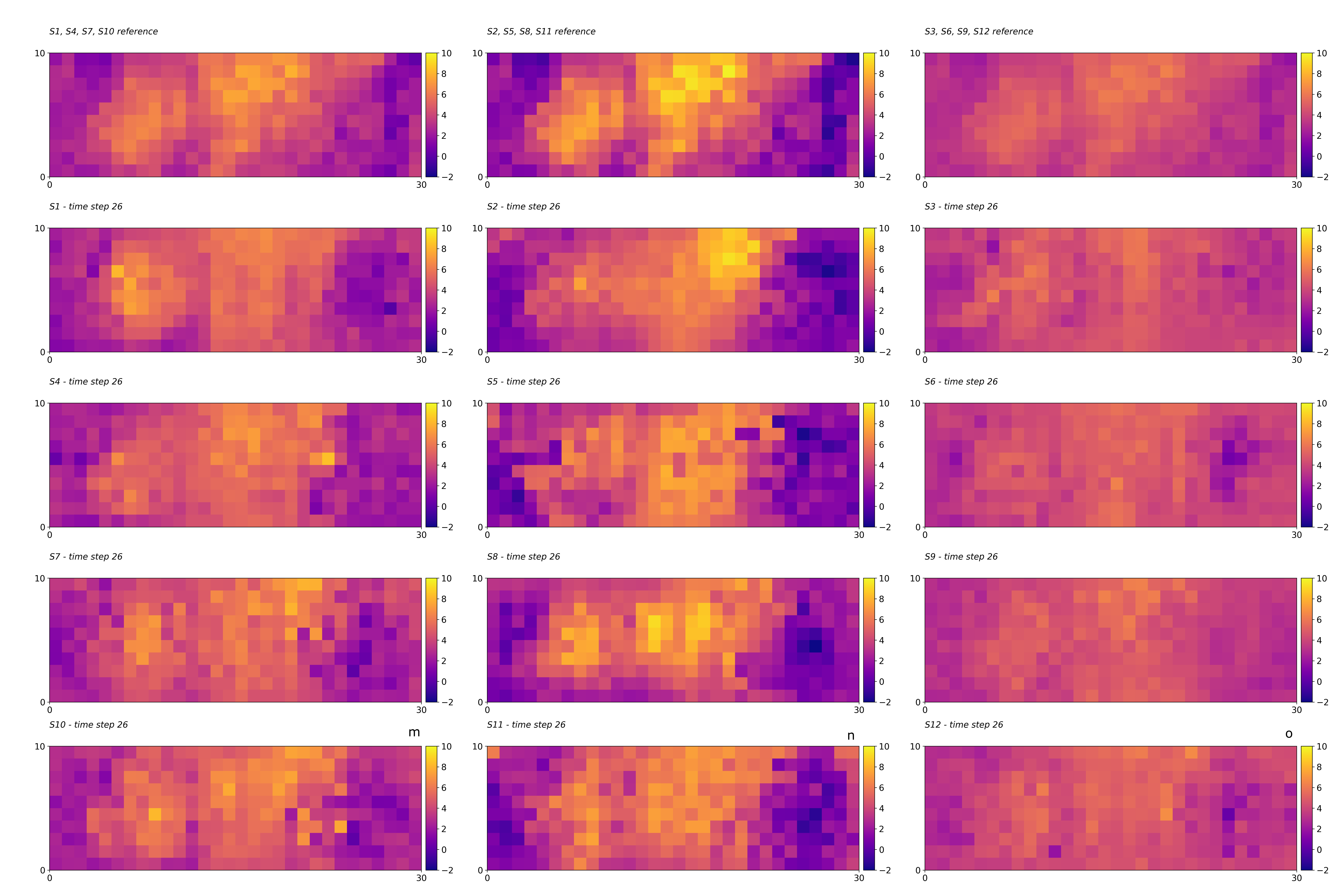}
\end{center}
\caption{Reference fields (top row) and final estimates of log-conductivity by the mean of the ensemble at time step 26 for the different scenarios.}
\label{fig:Figure 6}
\end{figure}

\begin{figure}[htbp]
\begin{center}
\includegraphics*[scale=0.3, angle=0]{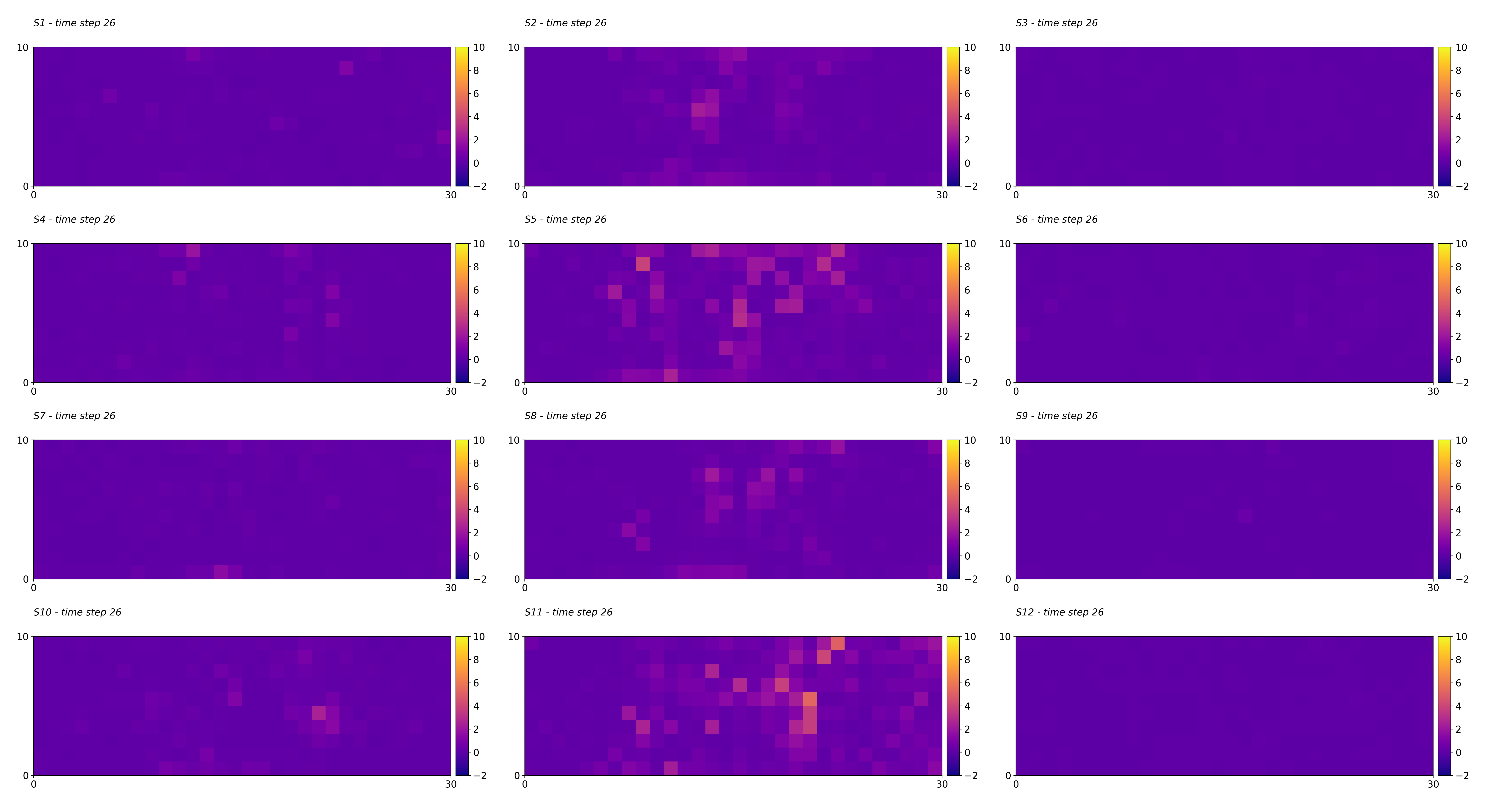}
\end{center}
\caption{Variance of the final ensemble of realizations for the different scenarios.}
\label{fig:Figure 7}
\end{figure}

\begin{figure}[htbp]
\begin{center}
\includegraphics*[scale=0.30, angle=0]{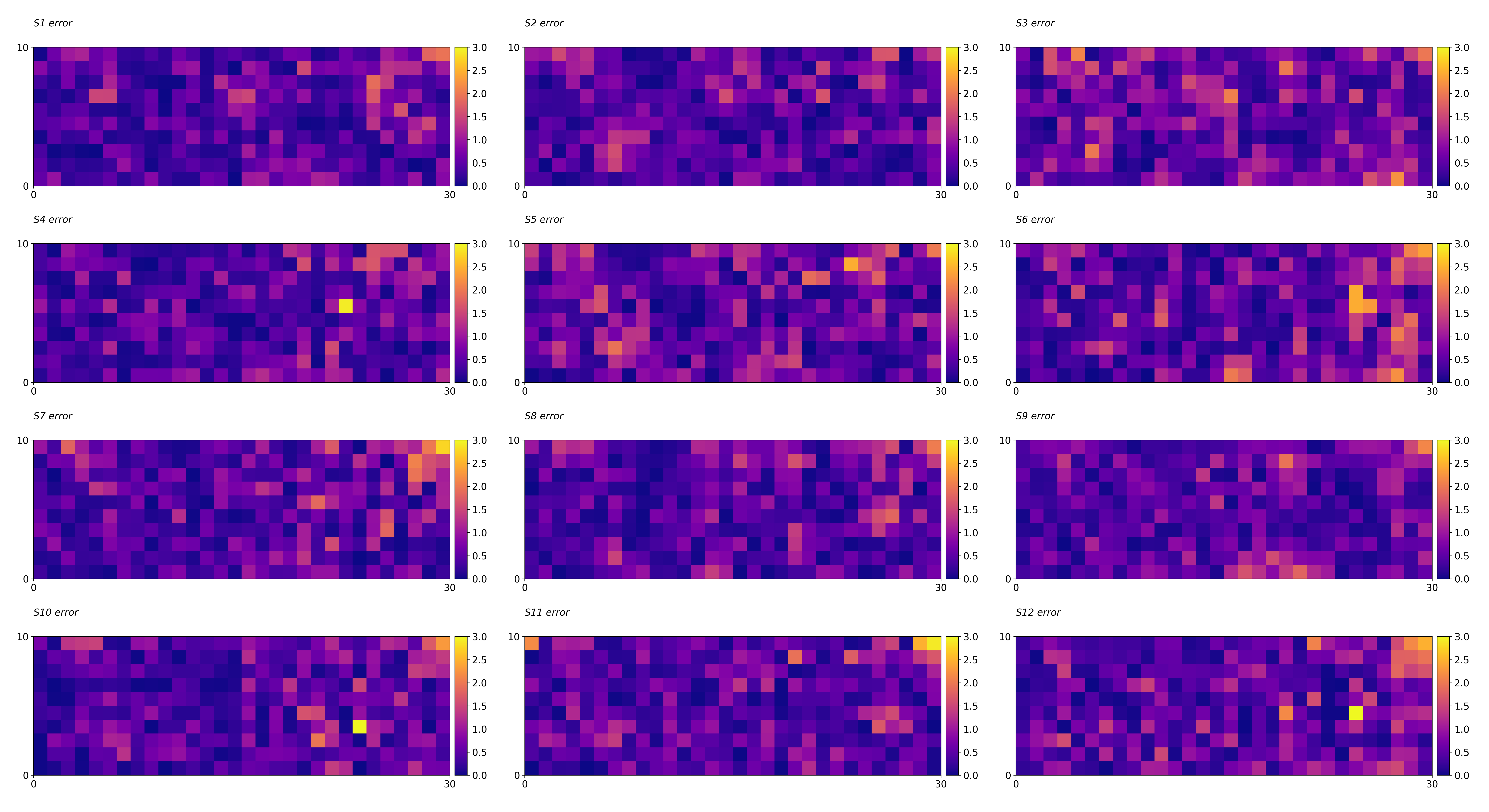}
\end{center}
\caption{Standardized deviations between the reference fields and the mean of the final ensemble of realizations for the different scenarios.}
\label{fig:Figure 8}
\end{figure}
\end{landscape}

\begin{figure}[htbp]
\begin{center}
\includegraphics*[scale=0.3, angle=0]{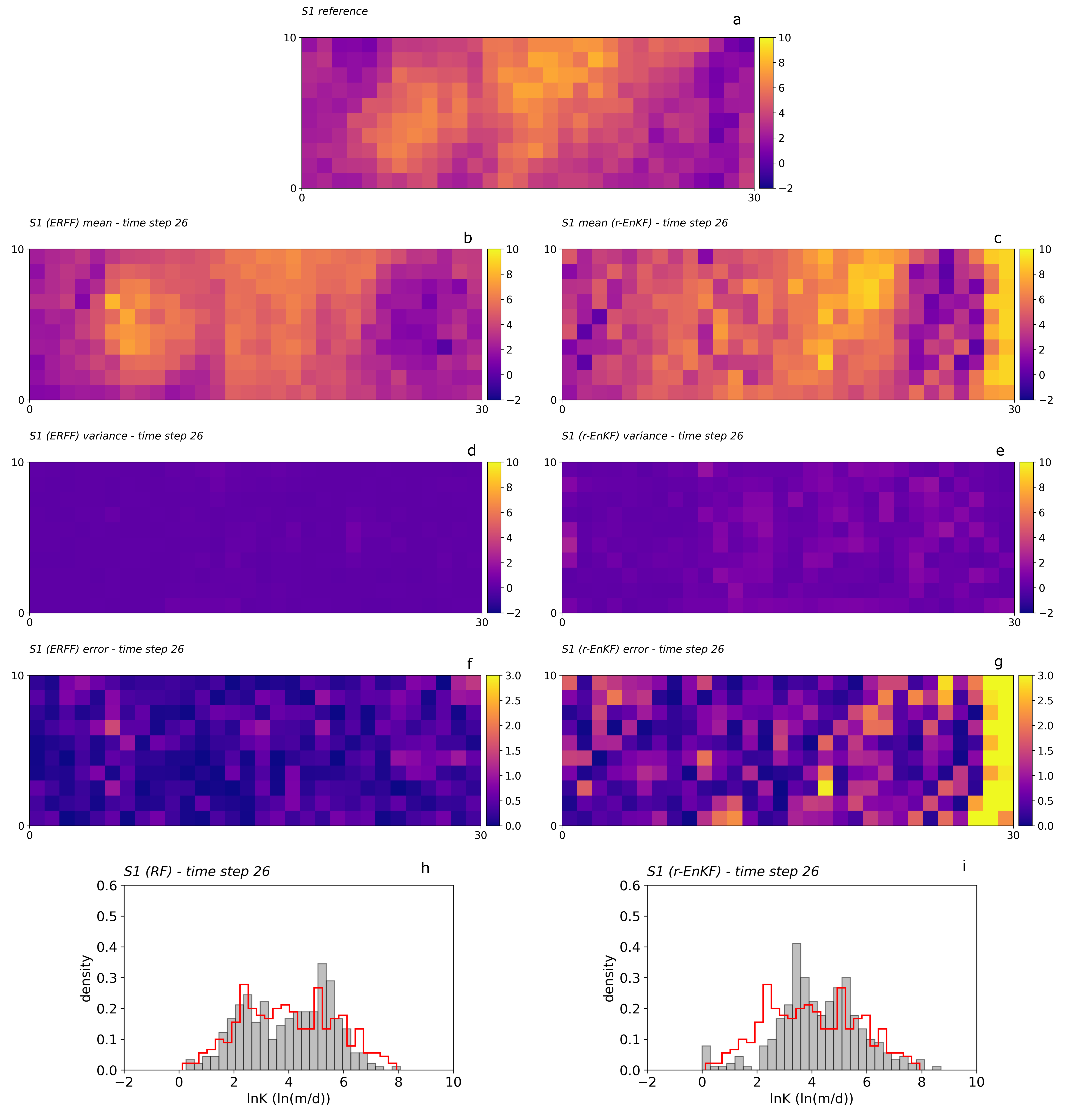}
\end{center}
\caption{Comparison between the ensemble random forest filter and the restart ensemble Kalman filter for scenario S1. Reference field, mean and variance of the ensemble of realizations, and mean field histogram.}
\label{fig:Figure 9}
\end{figure}

Aware of the very good results that the r-EnKF had given in the past, the exercise was repeated with an ensemble of 500 realizations, and then it was able to yield results as good as the ERFF but with a computational cost three times larger.  

\begin{figure}[htbp]
\begin{center}
\includegraphics*[scale=0.50, angle=0]{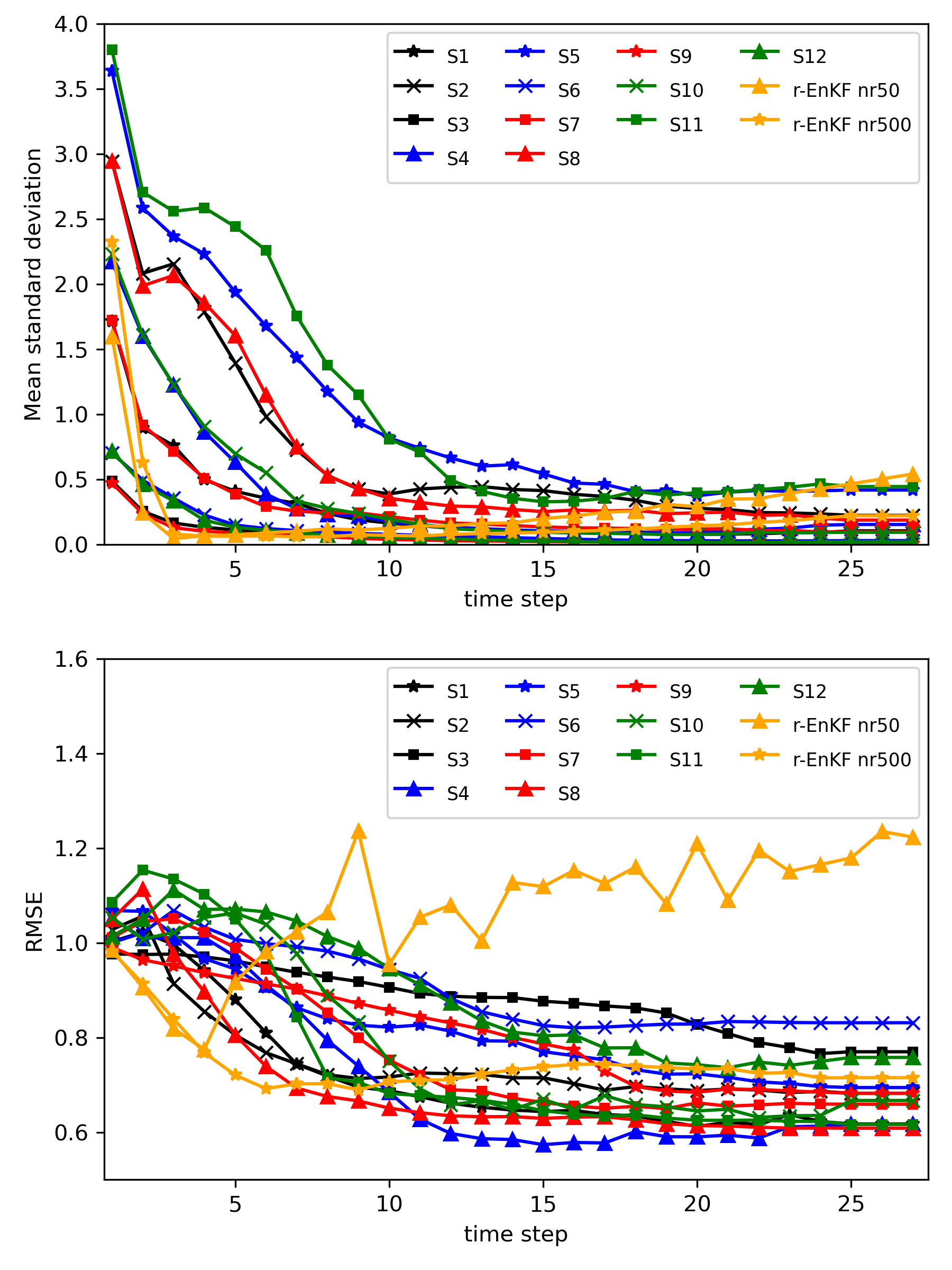}
\end{center}
\caption{ASD and RMSE}
\label{fig:Figure 10}
\end{figure}

For quantitative analysis, the root-mean-square errors (RMSE) and the average standard deviations (ASD) were computed according to 

\begin{eqnarray}
	RMSE &=&\sqrt{\frac{1}{n_e n_p}\sum_{i=1}^{n_p}\sum_{j=1}^{n_e}{(x_{ij}-x_i^{ref})^2}} \\
	ASD &=& \frac{1}{n_p}\sum_{i=1}^{n_p}	{\sigma_{x_i}} 
\end{eqnarray}
where $n_e$ is the number of realizations in the ensemble, $n_p$ is the number of cells, $x_{ij}$ represents the log-conductivity at cell $i$ in realization $j$, $x_i^{ref}$ is the log-conductivity in the reference, and $\sigma_{x_i}$ is the log-conductivity ensemble standard deviation at cell $i$. Their evolution in time is shown in Figure~\ref{fig:Figure 10}, for all 12 scenarios plus the r-EnKF with 50 and 500 ensemble realizations. Both values decrease in magnitude as time passes with the best performer being S11 (highest values for number of observations, number of realizations, and also reference variance) followed by S8 (same as S11 but with only 50 realizations). Note also, how the RMSE goes chaotic for the r-EnKF with 50 realizations after iteration 5, probably due to a problem with filter inbreeding, very common in ensemble Kalman filter analyses. 

\begin{figure}[htbp]
\begin{center}
\includegraphics*[scale=0.60, angle=0]{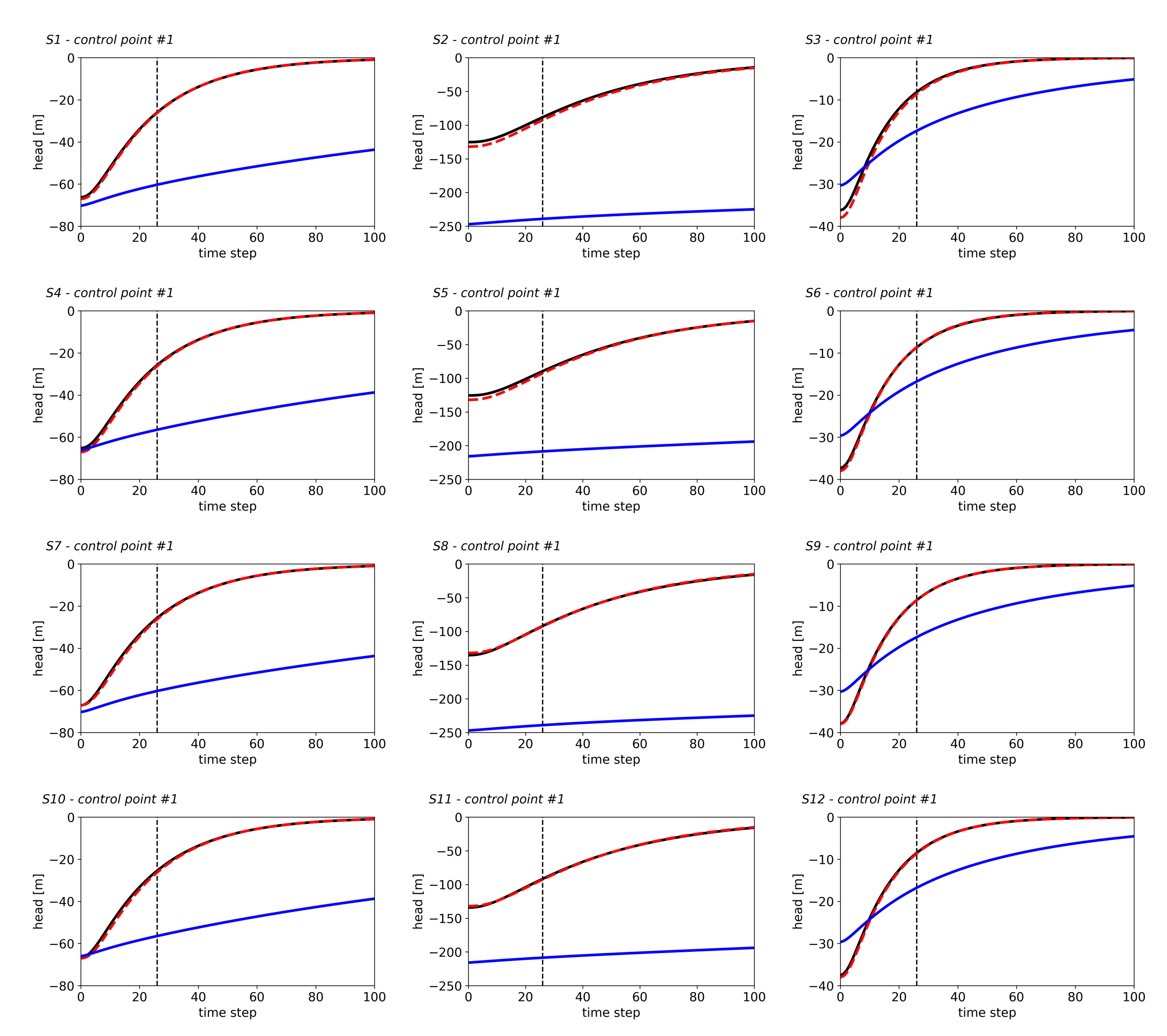}
\end{center}
\caption{Head evolution at control point \#1. Reference field (dashed line). Mean of head simulations in the initial log-conductivity ensembles (blue solid line). Mean of head simulations in the final ensembles (black solid line).}
\label{fig:Figure 11}
\end{figure}

\begin{figure}[htbp]
\begin{center}
\includegraphics*[scale=0.60, angle=0]{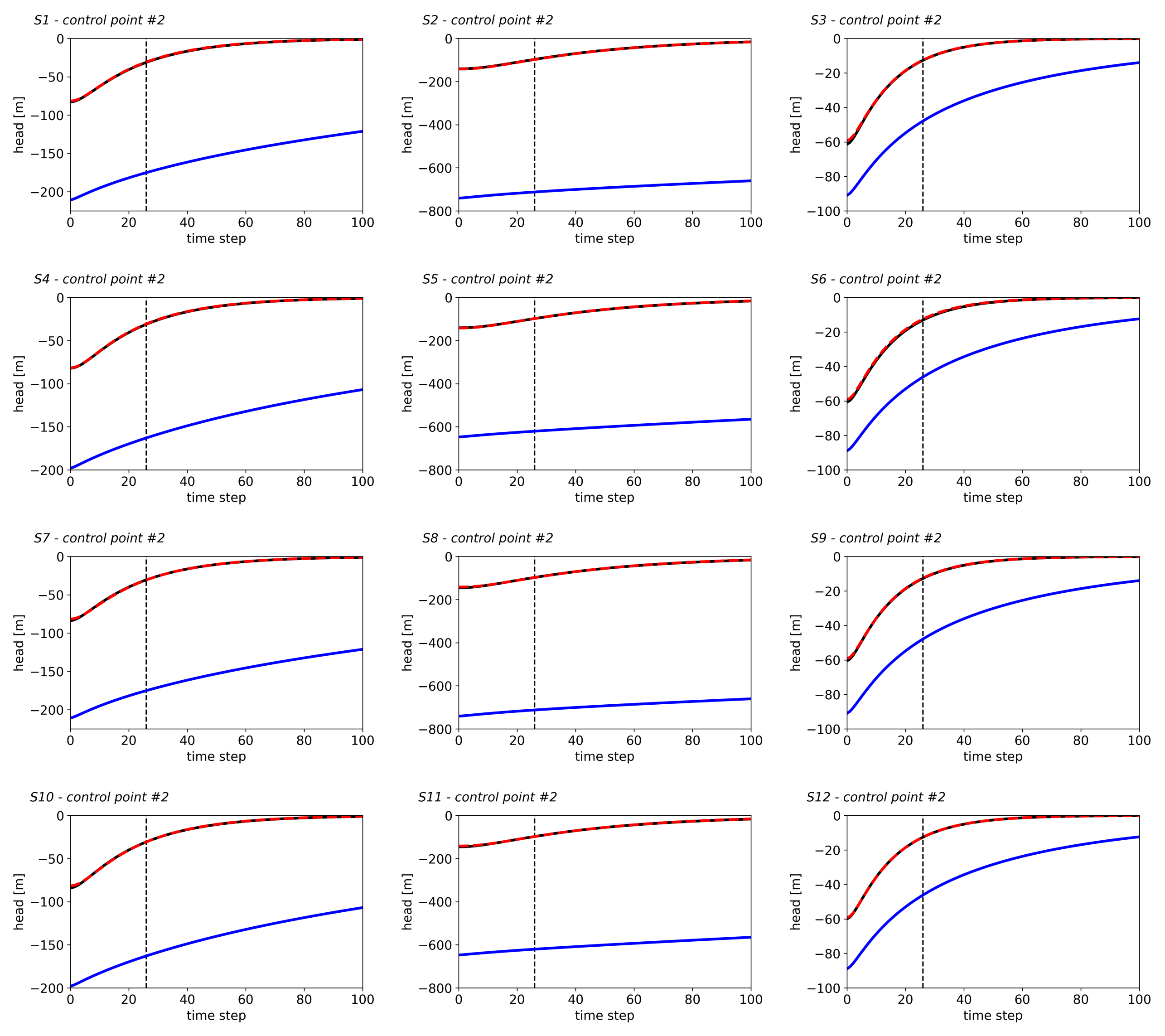}
\end{center}
\caption{Head evolution at control point \#2. Reference field (dashed line). Mean of head simulations in the initial log-conductivity ensembles (blue solid line). Mean of head simulations in the final ensembles (black solid line).}
\label{fig:Figure 12}
\end{figure}

\begin{figure}[htbp]
\begin{center}
\includegraphics*[scale=0.6, angle=0]{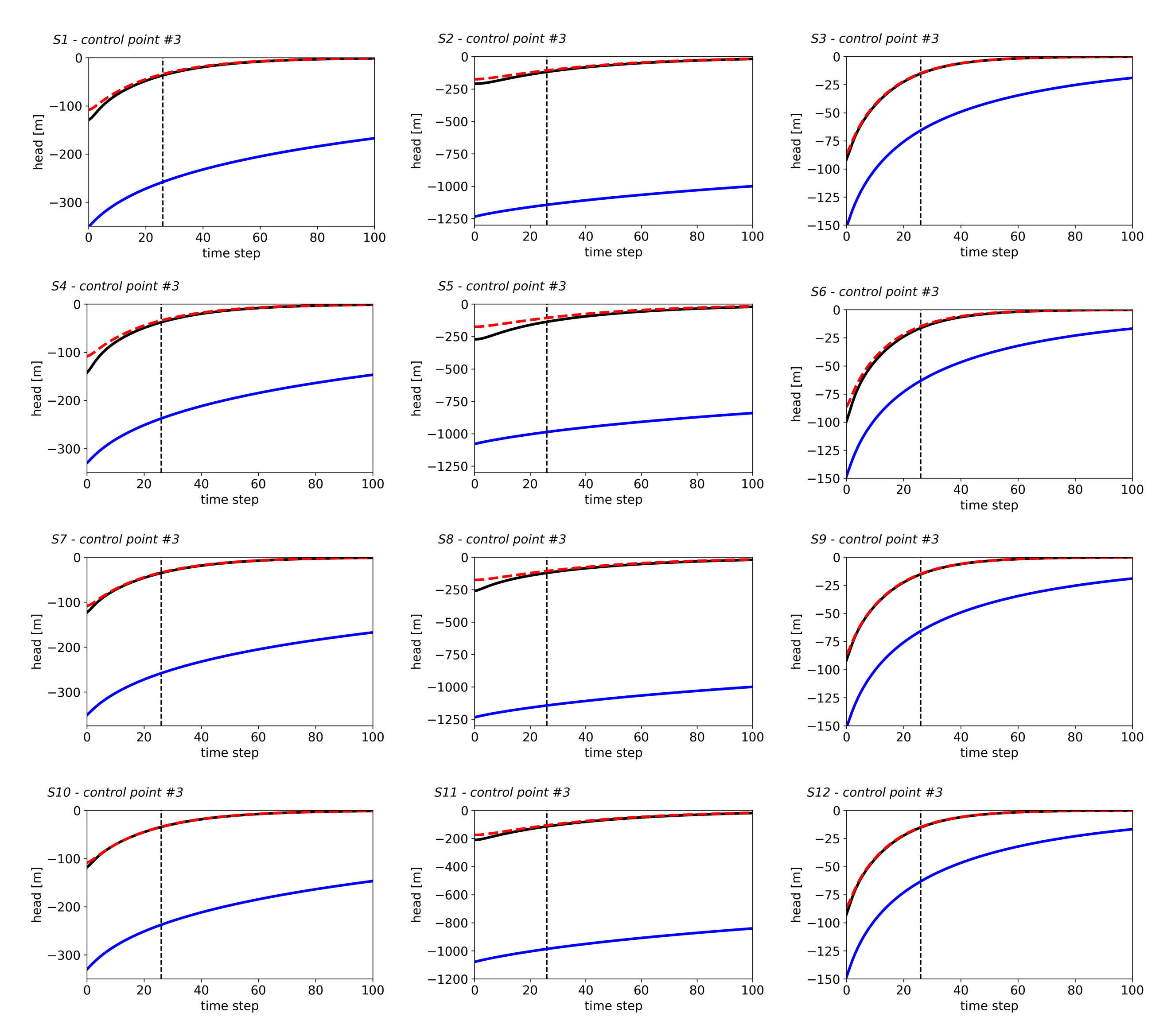}
\end{center}
\caption{Head evolution at control point \#3. Reference field (dashed line). Mean of head simulations in the initial log-conductivity ensembles (blue solid line). Mean of head simulations in the final ensembles (black solid line).}
\label{fig:Figure 13}
\end{figure}

\begin{figure}[htbp]
\begin{center}
\includegraphics*[scale=0.70, angle=0]{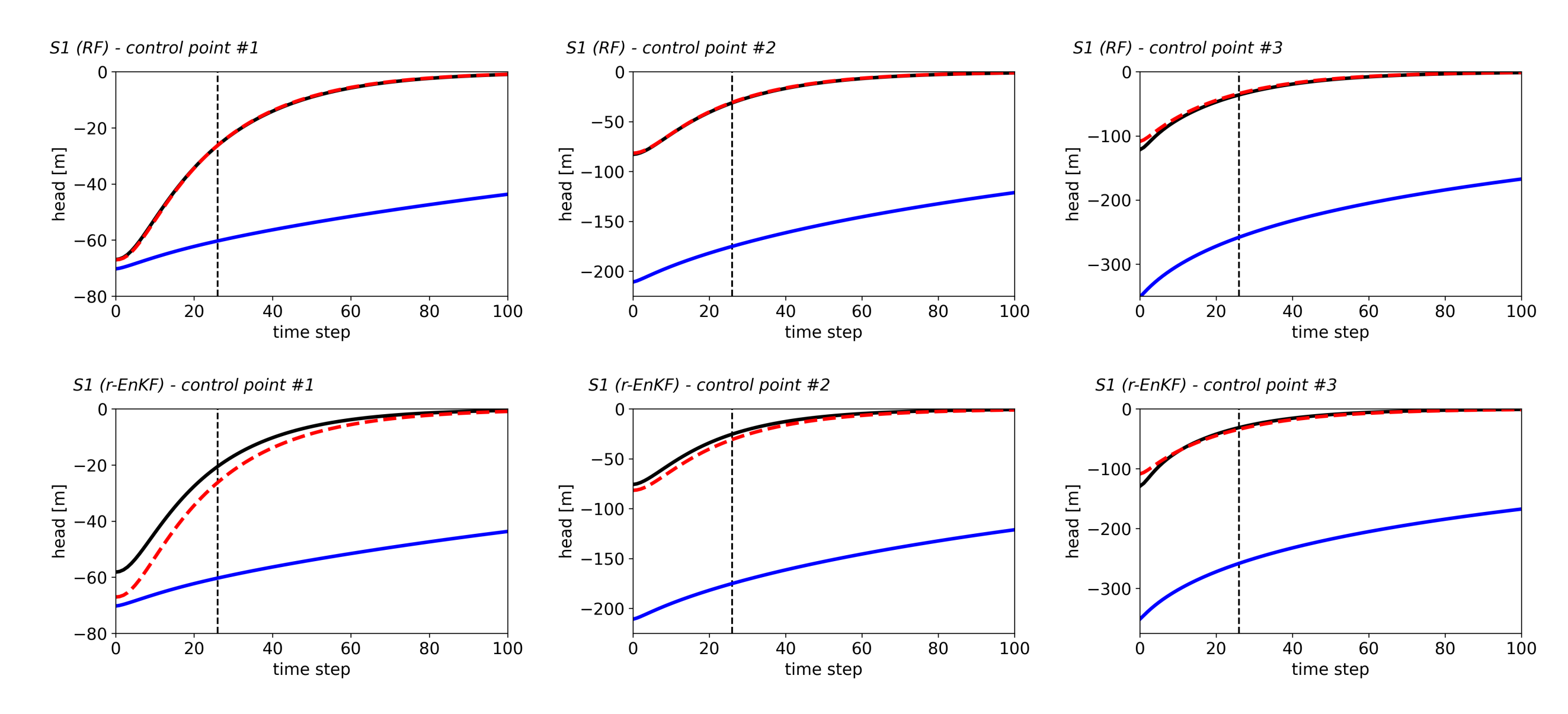}
\end{center}
\caption{Head evolution at the three controlo points for scenario S1 in the final ensembles of realizations obtained by the ERFF and by the r-EnKF with 50 realizations. Reference field (dashed line). Mean of head simulations in the initial log-conductivity ensembles (blue solid line). Mean of head simulations in the final ensembles (black solid line).}
\label{fig:Figure 14}
\end{figure}

Finally, Figure \ref{fig:Figure 11}, Figure \ref{fig:Figure 12}, and Figure \ref{fig:Figure 13} show how the piezometric heads are reproduced at the three control points. Observations were assimilated only until time step 26 (vertical dashed line in all plots), but the piezometric head evolution is shown until the end of the simulation period at time 100. All figures show the head simulation in the reference field from time zero (dashed red line), the average of all head simulations in the initial ensemble of realizations (solid blue line), and the average of the simulations in the updated log-conductivity fields after 26 assimilation steps (solid black line). Note that to best display the results the piezometric head axes vary for each plot. The graphs have been grouped by columns, with each column corresponding to one of the three reference cases. It is quite remarkable how the piezometric heads change from being completely off target at time zero to matching, almost perfectly, the reference head curves. The minimal discrepancies between the mean of the simulated values and the reference happen in some of the scenarios with the smaller number of observations, i.e., S4 and S5. For comparison purposes, the head evolution in the log-conductivity realizations obtained using the r-EnKF with 50 realizations is shown in Figure~\ref{fig:Figure 14}, where it can be seen that the reproduction of the reference values is not as good as for the ERFF, particularly for control points \# 1 and \# 2. 

\section{Conclusion}

A new data assimilation method, the ensemble random forest filter (ERFF), has been proposed. It is inspired by the ensemble Kalman filter but replaces the linear updating step with a non-linear update computed using random forests. The ERFF takes advantage of the use of an ensemble of log-conductivity realizations and its associated ensemble of predicted piezometric heads to build a large training dataset from few realizations (the dataset size grows with the square of the number of realizations). The random forest analyzes the differences in the predicted piezometric heads at observation locations with the differences in log-conductivities throughout the domain, learns from this training set, and then predicts what should be the difference to be added to the log-conductivity at each location in the domain once the head observations are collected and their differences with respect to the predictions evaluated. 

The method has been tested in a number of scenarios with varying degrees of heterogeneity (as measured by the standard deviation), different number of realizations in the ensemble, and different number of observation locations, and it has been found to perform well in all scenarios and better than its benchmarking the restart ensemble Kalman filter when the same number of realizations are used. Only when the number of realizations rises to 500, is the Kalman filter capable of providing similar results but at a cost three times larger than the ERFF. 

The main caveat of the proposal is, as in most machine learning applications, the choice of the hyperparameters that control the building of the random forests. This task could be time-consuming until a suitable set of hyperparameters is found that performs appropriately for the problem at hand. 

Research continues on the application of the ERFF to more complex problems, such as those involving the identification of external stresses and boundary and initial conditions or the identification of more complex log-conductivity patterns. 

\section*{Conflict of Interest Statement}

The authors declare that the research was conducted in the absence of any commercial or financial relationships that could be construed as a potential conflict of interest.

\section*{CRediT authorship contribution statement}

{\bf Vanessa A. Godoy:} Methodology, Software,  Validation, Formal analysis, Investigation, Writing - Original Draft, Visualization 
{\bf Gian Napa-Garc\'ia:} Methodology, Software
{\bf J. Jaime G\'omez-Hern\'andez:} Methodology, Software, Validation, Formal analysis, Investigation, Writing - Review \& Editing, Visualization, Supervision, Funding acquisition. 

\section*{Funding}
The authors acknowledge grant PID2019-109131RB-I00 funded by \\ MCIN/AEI/10.13039/501100011033 and project InTheMED, which is part of the PRIMA Programme supported by the European Union's Horizon 2020 Research and Innovation Programme under Grant Agreement No 1923.

\section*{Data Availability Statement}
The datasets and code used are available under reasonable request from the corresponding author.

\bibliographystyle{acm}
\bibliography{ML_normal_bib} 

\end{document}